\documentclass[letterpaper]{article}
\usepackage{iccc}

\usepackage{times}
\usepackage{helvet}
\usepackage{courier}
\usepackage{booktabs}
\usepackage{subcaption}
\usepackage{caption}
\usepackage{graphicx}
\usepackage{color}
\usepackage{amsmath}
\usepackage[ruled]{algorithm2e}
\usepackage[dvipsnames]{xcolor}

\title{Visual Conceptual Blending with Large-scale Language and Vision Models}
\author{Songwei Ge$^{1}$ and Devi Parikh$^{2}$\\
$^{1}$University of Maryland \\ $^{2}$Facebook AI Research \& Georgia Tech\\
}

\begin{document} 

\maketitle

\begin{abstract}
\begin{quote}
We ask the question: to what extent can recent large-scale language and image generation models blend visual concepts? Given an arbitrary object, we identify a relevant object and generate a single-sentence  description of the blend of the two using a language model. We then generate a visual depiction of the blend using a text-based image generation model. Quantitative and qualitative evaluations demonstrate the superiority of language models over classical methods for conceptual blending, and of recent large-scale image generation models over prior models for the visual depiction.
\end{quote}
\end{abstract}

\section{Introduction}

Throughout the development of human civilization, our unique capacity to blend unfamiliar concepts has led to innovation of advanced tools, invention of new art styles, and breakthroughs in science. Machines demonstrating this ability is considered to be one of the hallmarks of creativity and intelligence. Such  systems could help understand human creativity. Moreover, they can assist humans in exploring the inexhaustible space of combinations of different concepts. This has been an area of research for decades ~\cite{fauconnier1998conceptual}, which has led to both theoretical work~\cite{cunha2020let} as well as prototypes of support tools to assist users~\cite{karimi2018computational,chilton2019visiblends}. In the meantime, deep learning has achieved exceptional success in many areas where humans  excelled, from beating the best professional player in Go to making creative advertising designs.

\begin{figure}[t]
    \centering
    \begin{subfigure}{0.48\linewidth}
        \includegraphics[trim=0 0 0 0,clip,width=\linewidth]{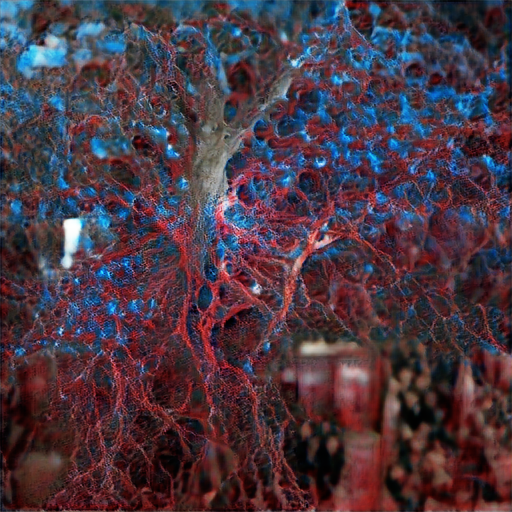}
        \caption{``A tree made of blue and red blood vessels''.}
    \end{subfigure}
    \hfill
    \begin{subfigure}{0.48\linewidth}
        \includegraphics[trim=0 0 0 0,clip,width=\linewidth]{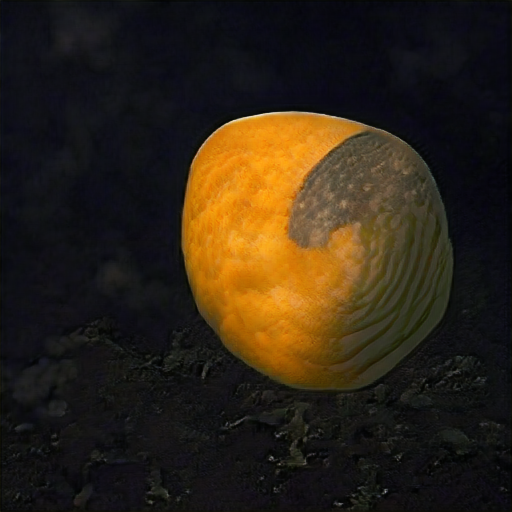}
        \caption{``The moon that is sliced like an orange''.}
    \end{subfigure}
    \vspace{-0.2cm}
    \caption{Visual conceptual blends generated by our framework using large-scale  language and vision models.}
    % \devi{These don't look very nice to me :) Should we show just 2 examples say, and use some of the better ones we had seen (e.g., on the Slack channel when we first tried these things)?} }
    \vspace{-0.5cm}
    \label{fig:teaser}
\end{figure}

In this paper, we examine deep neural networks trained on large-scale data in a general scenario of visual conceptual blending: given a single object as input (e.g., moon), can a relevant object be identified (e.g., an orange), can a relevant property that a blend can hinge on be identified (e.g., sliced), and finally, can an image be generated to depict the blend (e.g., ``the moon sliced like an orange'')? We use prompt-engineering with language models for the reasoning phase (identifying a relevant object and property), and text-based image generation models for the visualization phase. See Figure~\ref{fig:teaser} for example outputs.

We compare our approach quantitatively and qualitatively to representative existing approaches. To evaluate the ability to associate concepts, we compare our approach to traditional knowledge bases on a simile dataset. To evaluate the visual generation, we compare our approach to an existing GAN approach via human studies.
We show that large-scale models significantly outperform these baseline models. In general, we find that an appropriate composition of recent large-scale models results in encouraging creative abilities like visual conceptual blending.

\section{Related Work}

\vspace{3pt}
\noindent \textbf{Visual Conceptual Blending} Fauconnier and Turner first proposed the idea of conceptual blending and pointed out its indispensability in human development~\cite{fauconnier1998conceptual,fauconnier2008way}. Cognitive and neural scientists have been fascinated by the human ability to blend concepts and view such an ability as a milestone for AI development~\cite{eppe2018computational}.  More practically, the idea of visual conceptual blending has been applied in many commercial areas from advertising, journalism, to public service announcements~\cite{chilton2019visiblends}. In this section, we discuss the recent progress in developing systems that automatically blend visual concepts and the studies that measure the success of conceptual blending. 

Many systems developed by these studies act as support tools for augmenting human creativity.  \cite{chilton2019visiblends} presents a workflow where users identify the associated concepts, retrieve appropriate images, and label the analogous parts of the objects while the system automatically blends the images by combining these common parts. \cite{karimi2018computational}  explores visual conceptual blends in the context of sketching by leveraging the idea of concept shifts. \cite{cunha2017pig} proposes a description-based method that can blend sketches using detailed annotations. See \cite{cunha2020let} for a road map of visual conceptual blending.
\cite{mccaig2016deep,berov2016visual} applies style transfer models and the deep dream algorithm to render an image in a particular artistic style. 
\cite{Sbai_2021_CVPR} studies placing objects in uncommon contexts using a search-and-compose method.
Measuring the creativity of visual blends is known to be difficult. Fauconnier and Turner proposed several optimality principles to guide the conceptual blending~\cite{fauconnier1998conceptual}.  \cite{martins2015good} analyzes what makes a good blend using 15 hybrid animal images and a questionnaire.

\vspace{3pt}
\noindent \textbf{Analogical Reasoning with Language Models} Language models were first proposed to model the sequential nature of  language~\cite{mikolov2012context}. With the increasing sizes of training data and model capacities,  large-scale language models such as BERT~\cite{devlin2018bert} fine-tuned on the downstream tasks have dominated standard leaderboards.
Interestingly, several recent studies use language models as  knowledge bases to solve different problems without training on the task of interest~\cite{petroni2019language,jiang2020can}. These methods rely on task-specific prompts -- converting the task of interest to that of language modeling. Letting the language model predict masked parts from the prompt then becomes equivalent to the model solving the task of interest~\cite{petroni2019language,jiang2020can}. 
We propose to apply a similar idea to concept blending -- we design appropriate prompts to identify relevant concepts and properties along which to blend the concepts. Analogical reasoning has also been approached with large-scale knowledge bases~\cite{liu2017analogical}.
However, knowledge bases are known to be incomplete and rigid. We argue that this makes them less suitable for associating concepts in flexible ways ~\cite{cunha2020let}. 

\vspace{3pt}
\noindent \textbf{Deep Generative Models for Images} Most state-of-the-art image generation methods are built on either Generative Adversarial Networks (GANs)~\cite{goodfellow2014generative} or Variational AutoEncoders (VAEs)~\cite{kingma2013auto}. 
In this paper, we are primarily interested in generating conceptually blended objects. \cite{bau2020rewriting} proposes to modify the images through manipulating the intermediate layers in GANs which admits the possibility to blend concepts. In this work we use a textual description of the blend to guide the generation. %The most relevant research area to our problem is 
Text-based image generation models~\cite{reed2016generative,Zhu_2019_CVPR,tao2020df} are relevant.
DALL·E~\cite{ramesh2021zeroshot} is one such recent model that uses a pretrained discrete VAE  to compress images into low-dimensional vectors and then models the joint distribution of the vectors with text embeddings autoregressively.

\section{Approach}
Next, we describe how we use large language and image generation models to produce conceptually blended images given an input object. We decompose the visual conceptual blending process into two phases: reasoning and generation.

\vspace{3pt}
\noindent \textbf{Reasoning Phase}
The reasoning phase produces a textual description of the blend. 
We formulate the problem as follows: given an input object, the model identifies a relevant object and generates a description of the blend of the two. Note that our setting is more general than one where both concepts to be blended are given as input~\cite{cunha2020let}. 
We use \textit{moon} as the example input to explain the details of our prompt engineering approach.

To identify a relevant object, we use a simile-inducing input: ``the moon is like a [MASK]” and ask the language model to predict the masked word. The language model produces \textit{ghost}, i.e. ``the moon is like a ghost”. Next, we utilize the prompt ``the ghost has the property of [MASK]'', where the language model predicts the word \textit{dead}. We plug the predictions into a template and produce the description of the blend ``a moon that is dead like a ghost”. Other concepts and their properties identified using a pretrained BERT~\cite{devlin2018bert} model are shown in Table~\ref{tab:trial1}. Sometimes the retrieved objects are semantically similar rather than visually similar to the \textit{moon} such as \textit{ghost} and \textit{dream}. We see some interesting blends such as ``a moon that is lit like a beacon" and ``a moon that is broken like a rainbow".
\begin{table}
\small
\centering
\caption{Top 5 concepts relevant to \emph{moon}, and associated properties using simile-inducing prompts to a BERT model.}
\vspace{-0.1cm}
\label{tab:trial1}
\begin{tabular}{l|lllll}
\toprule
concept & \multicolumn{5}{c}{property} \\
\midrule
ghost   & dead      & killed  & gone     & alive     & murdered \\
dream   & over      & real    & complete    & gone     & broken  \\
rainbow & broken      & colorful     & green    & white     & black    \\
beacon  & lit       & active  & red      & closed    & automated   \\
jewel   & lost      & gone    & precious & beautiful & gold     \\
\bottomrule
\end{tabular}
\vspace{-0.1cm}
\end{table}

\begin{table}
\small
\centering
\caption{Top 5 concepts relevant \emph{moon}, and associated properties using shape-guided prompts to a BERT model.}
\vspace{-0.1cm}
\label{tab:trial2}
\begin{tabular}{l|lllll}
\toprule
concept & \multicolumn{5}{c}{property} \\
\midrule
shell  & white     & smooth     & thin   & brown & small   \\
head   & rounded   & black      & white  & brown & small   \\
fruit  & edible    & white      & yellow & red   & purple  \\
egg    & white     & yellow     & laid   & blue  & red     \\
eye    & open      & closed     & small  & red   & black   \\
\bottomrule
\end{tabular}
\vspace{-0.1cm}
\end{table}

\begin{figure}[t]
    \centering
    \includegraphics[trim=0 5 0 0,clip,width=\linewidth]{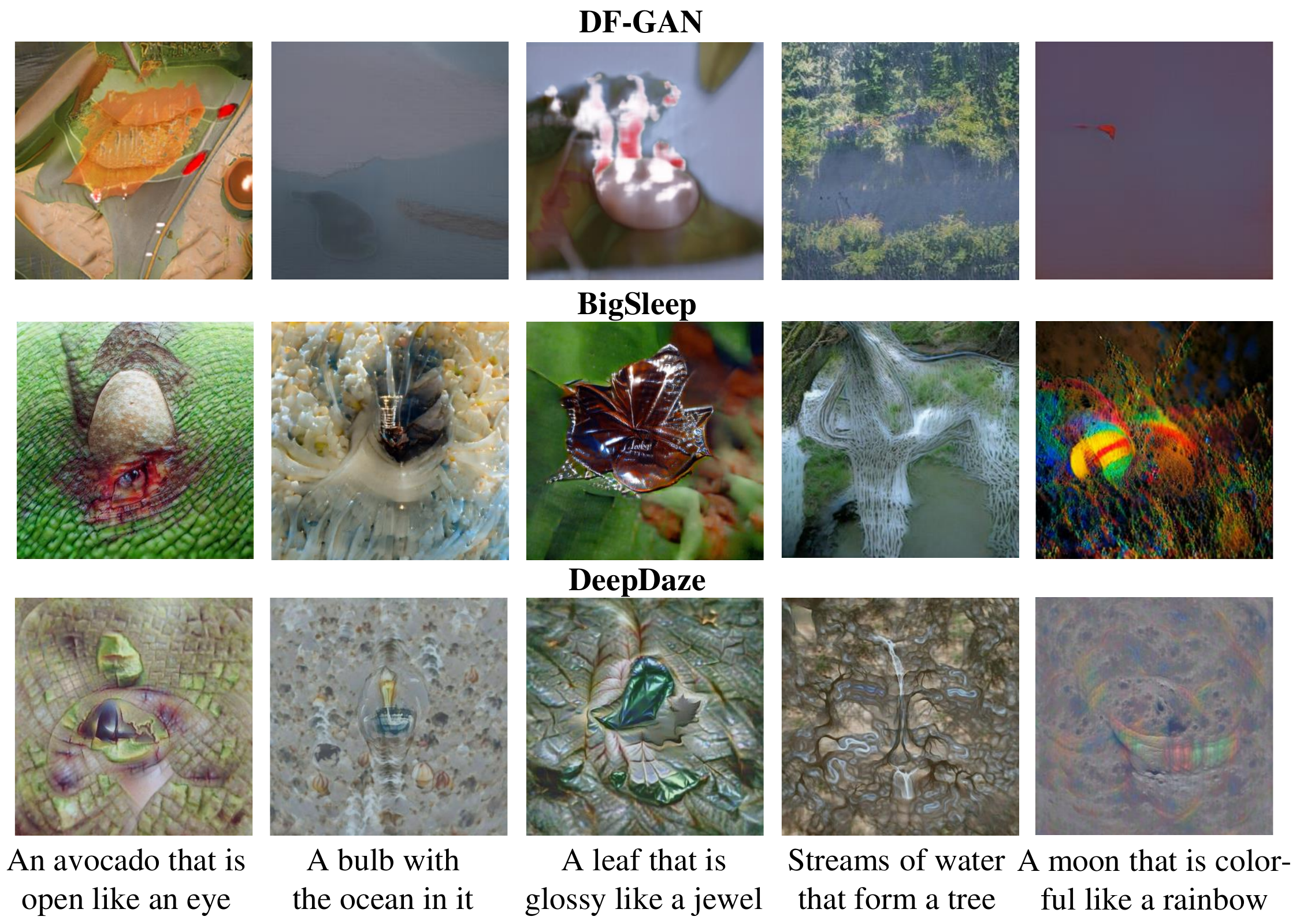}
    \caption{Visual blends generated using different methods using blend descriptions shown at the bottom as input.}
    \label{fig:result}
    \vspace{-0.5cm}
\end{figure}

Shape is often recognized as the bridge to connect different visual concepts~\cite{steinbruck2013conceptual,chilton2019visiblends}. This motivates a shape-guided prompt to identify relevant objects. Specifically, we first use language models to predict the shape of the \textit{moon} with the prompt ``The shape of the moon is [MASK1]”. The language model outputs \textit{spherical}. This gives us ``The shape of the moon is spherical”. Then we plug the word \textit{spherical} into the prompt ``The shape of the [MASK2] is  spherical'', and the language model predicts the relevant object \textit{shell}, i.e. ``The shape of the  shell is spherical". This leads to a blend description ``a moon that is smooth like a shell'' with the property \textit{smooth} of the \textit{shell}. More identified concepts and their properties are shown in Table~\ref{tab:trial2}. We find that the candidate concepts we obtain are visually similar to the \textit{moon} in terms of shape. Some interesting descriptions include ``a moon that is laid like a egg" and ``a moon that is edible like a fruit". In practice, shape can be replaced by other properties that connect visual concepts. For example, speed connects \textit{bullet} and \textit{runner} and reflection connects \textit{mirror} and \textit{lake}.

\vspace{3pt}
\noindent \textbf{Generation Phase}
In this phase we generate an image based on the description output by the reasoning phase. To demonstrate the ability of large models in realizing the blends, we explore BigSleep\footnote{https://github.com/lucidrains/deep-daze} and DeepDaze\footnote{https://github.com/lucidrains/big-sleep} 
which utilize the CLIP model~\cite{unpublished2021clip} to guide the BigGAN~\cite{brock2019large} and SIREN~\cite{sitzmann2020implicit} models for text-based image generation. 

Specifically, suppose we are given a trained CLIP model $f_\theta(x_i, x_s)$ which takes an image $x_i$ and a sentence $x_s$ as input and outputs the similarity, and a trained BigGAN model $g_\phi(z)$ which takes a random Gaussian vector $z$ as input and outputs an image. We first sample a vector $z_0$ from a standard Gaussian distribution. $z$ is iteratively updated to maximize the similarity of the generation $ g_\phi(z_{t})$ and the text $x_s$ as computed by the CLIP model $f_\theta(x_i, x_s)$. 
DeepDaze adopts a similar process with BigGAN replaced by SIREN. 

Overall, we now have a full pipeline to go from an input concept (e.g., \textit{moon}) to a description of its blend with a related concept (e.g., ``a moon that is sliced like an orange'') to an image that depicts this blend (e.g., Figure~\ref{fig:teaser}).

\section{Evaluation}

\vspace{3pt}
\noindent \textbf{Reasoning Phase}
To evaluate how well language models blend concepts, we evaluate on the simile dataset~\cite{chakrabarty2020generating}. It contains pairs of literal input and its simile version in the form of $<$\textit{Source}, \textit{Target}$>$, e.g. $<$The city was beautiful, The city was like a painting$>$. It evaluates the model's ability to identify ``painting'' based on ``the beautiful city''. However, we found that the language is inconsistent across the dataset. For instance, many pairs lack a subject or use a pronoun as subject, e.g. $<$Felt worthless, Felt like a low budget film$>$. We instead focus our evaluation on the model's ability to accomplish the core reasoning step --  predicting the property ``worthless'' based on the object ``a low budget film''. Using heuristics for pre-processing, we extracted $66,442$ property-objects pairs for evaluation.

We compare language models to knowledge bases. For the language model we use the prompt ``a low budget film is [MASK]'' as the input and ask the model to generate candidate predictions for the masked word. We consider 4 trained language models: ELMO~\cite{Peters2018}, $\text{BERT}_{\text{Base}}$ and $\text{BERT}_{\text{Large}}$~\cite{devlin2018bert}, and GPT~\cite{radford2018improving}.  For knowledge base, we use ConceptNet~\cite{speer2017conceptnet} which contains relations including ``IsA'', ``HasA'', ``HasAProperty'', etc., which form candidate predictions for properties relevant to the object.

Note that sometimes the object in our dataset is described as a phrase including qualifiers (e.g., ``a low budget film'') while ConceptNet only contains the root objects. 
We use dependency parsing to find the root of the phrase and use it to query ConceptNet. In our example, ``film" instead of ``a low budget film" is used. After this processing, 96.34\% of objects from our evaluation set can be found in ConceptNet.

For each method, we produced 1000 candidates, and report the precision, i.e. percentage of time that the property (e.g., ``worthless'') is in the top 10, 100, 1000 candidates. 
 Note that the ConceptNet API does not offer a  straightforward way to request an exact number of relations for an object. Different objects have different number of properties associated with them. When requesting 1000 relations for objects in our evaluation set, $688.90$ were returned on average. As shown in Table~\ref{tab:lm_simile}, the precision using ConceptNet is significantly lower than using language models.

Additionally, we notice that using larger language models can further improve the precision. 
In general, these results demonstrate that language models are better at associating concepts than knowledge bases. We hypothesize this is due to their flexibility and comprehensiveness. 

\begin{table}[h]
\small
\centering
\caption{Precision of language models and knowledge base on the simile dataset.}
\vspace{-0.2cm}
\label{tab:lm_simile}
\begin{tabular}{l|lll}
\toprule
           & P@10 & P@100 & P@1000 \\
\midrule
ConceptNet &  1.12 & 2.70 & 5.90 \\ \hline
Elmo       & 0.13 & 7.69  & 37.33  \\
$\text{BERT}_{\text{Base}}$  & 1.59 & 15.72 & 53.08  \\
$\text{BERT}_{\text{Large}}$ & 1.42 & 15.89 & 46.56  \\
GPT        & 2.59 & 24.84 & 66.38 \\
\bottomrule
\end{tabular}
\end{table}

\begin{figure}[t]
    \centering
    \includegraphics[trim=70 20 70 0,clip,width=\linewidth]{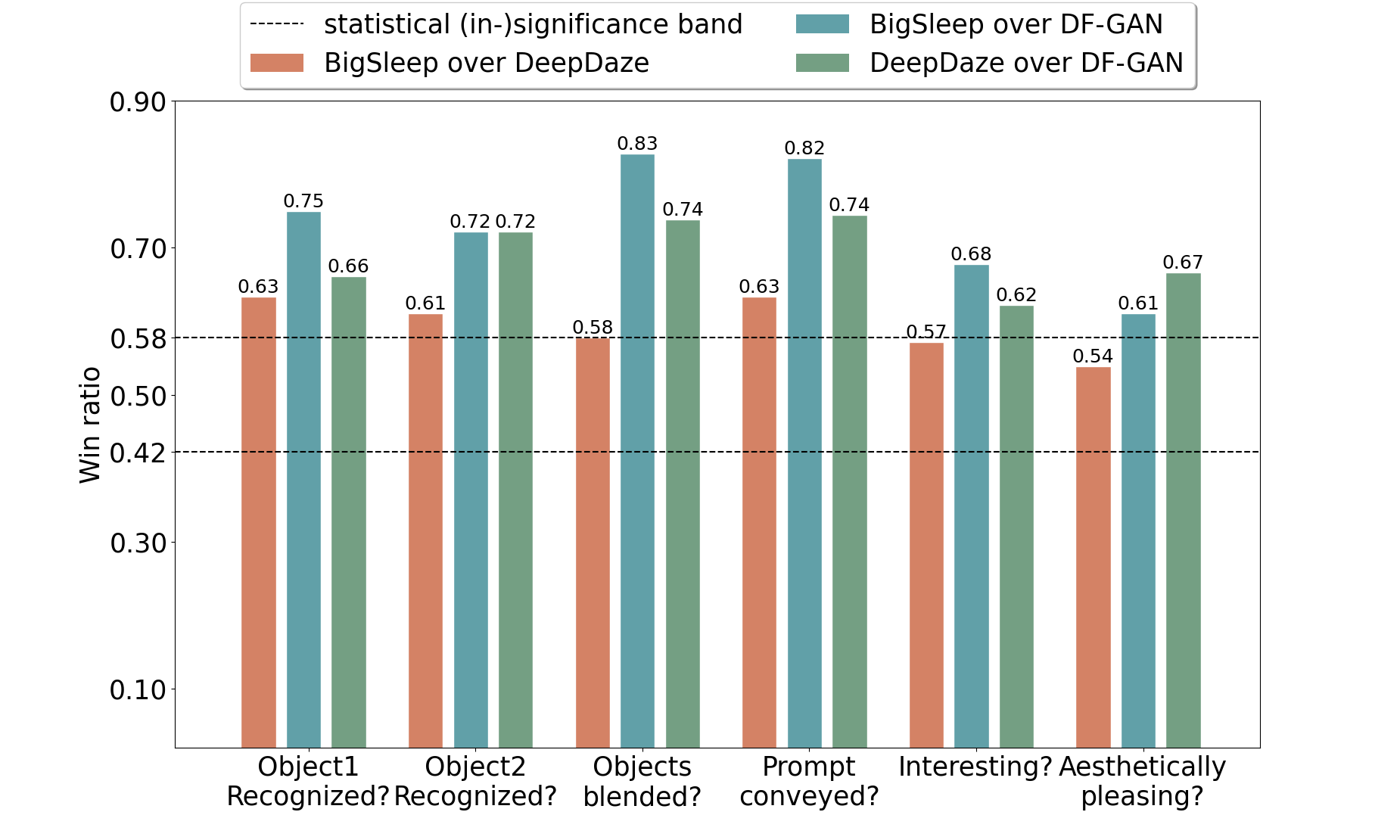}
    \caption{Human preference for different methods w.r.t. different questions. Values outside the band between the dashed lines are statistically significant at 95\% confidence.}
    \label{fig:human}
    \vspace{-0.5cm}
\end{figure}

\vspace{3pt}
\noindent \textbf{Generation Phase}
We collect 20 text descriptions of blends (see Figure~\ref{fig:result} for examples)
-- half generated with our reasoning approaches and rest by us. We use these descriptions as input to the large-scale BigSleep and DeepDaze models described earlier, as well as a recent DF-GAN~\cite{tao2020df} model. We run human evaluation on Amazon Mechanical Turk (AMT). We show subjects a pair of images generated by different methods to depict the visual blend of two objects and ask six questions: 
1. \textit{In which image do you recognize} OBJECT1 \textit{more?}   
2. \textit{In which image do you recognize} OBJECT2 \textit{more?}   
3. \textit{Which image blends the two objects better?}
4. \textit{Which image conveys the} DESCRIPTION \textit{better?}  
5. \textit{Which image looks more interesting to you?}
6. \textit{Which image looks more aesthetically pleasing to you?}  
These are designed using the optimality principles for concept blending~\cite{fauconnier1998conceptual}. Specifically, 1 and 2 relate to the unpacking principle, 3 and 4 to the integration principle, and 5 and 6 to general quality. Each question (6) for every pairwise comparison of models (3) and every textual description (20) is answered by $9$ unique subjects.

See results in Figure~\ref{fig:human}. The CLIP-based models (BigSleep and DeepDaze) significantly outperform DF-GAN, demonstrating the superiority of large models in generating visual blends. BigSleep is preferred over DeepDaze. We conjecture that this is because BigGAN learns a better prior on the image distribution than SIREN.

\section{Conclusion}
In this paper, we apply large-scale language and image generation models to a classic computational creativity problem -- visual conceptual blending. Our experiments show that these models allow us to use simple yet effective ways to generate visual blends that are significantly better than  previous methods. Future work includes engineering novel prompts to connect concepts and developing more complex blending strategies given the identified concepts. For example, the classic blend of boat and house (houseboat) -- ``a man lives in a house that is built on the water like a boat" -- considers structural relationships of the objects and includes two different properties from the two objects -- a place of accommodation (from house) and being on water (from boat).

{\small
\bibliographystyle{iccc}
\bibliography{iccc}
}

\end{document}